\def\year{2019}\relax
\documentclass[letterpaper]{article} 
\usepackage{aaai19}  
\usepackage{times}  
\usepackage{helvet}  
\usepackage{courier}  
\usepackage{url}  
\usepackage{graphicx}  

\usepackage{epsfig}
\usepackage{graphicx}
\usepackage{amsmath}
\usepackage{amssymb}
\usepackage{subfigure}
\usepackage{amsfonts}
\usepackage{setspace}
\usepackage{colortbl}
\usepackage{algpseudocode}
\usepackage{morefloats}
\usepackage[breaklinks=true,bookmarks=false]{hyperref}
\usepackage{morefloats}
\usepackage[boxed,lined,commentsnumbered]{algorithm2e}
\usepackage{relsize}
\usepackage{multirow}

\newcommand{\SH}{\textcolor{black}}
\newcommand{\ie}{\textit{i.e. }}
\newcommand{\eg}{\textit{e.g. }}

\newcommand{\FN}{\textcolor{black}}

\begin{document}
%

\title{Object Detection based on  Region Decomposition and Assembly}


\author{Seung-Hwan Bae\\
Computer Vision and Learning Lab., Department of Computer Engineering \\
Inha University, 100 Inha-ro, Michuhol-gu, Incheon, 22212, South Korea\\
shbae@inha.ac.kr\\
}

\maketitle
\begin{abstract}
Region-based object detection infers object regions for one or more categories in an image. Due to the recent advances in deep learning and region proposal methods, object detectors based on convolutional neural networks (CNNs) have been flourishing and  provided the promising detection results. However, the detection accuracy is degraded often because of the low discriminability of object CNN features caused by occlusions and inaccurate region proposals. In this paper, we therefore propose a region decomposition and assembly detector (R-DAD) for more accurate object detection.

In the proposed R-DAD, we first decompose an object region into multiple small regions. To capture an entire appearance and part details of the object jointly, we  extract  CNN features within the whole object region and decomposed regions. We then learn the semantic relations between the object and its parts by combining the multi-region features stage by stage with region assembly blocks, and use the combined and high-level semantic features for the object classification and localization. In addition, for more accurate region proposals, we propose a multi-scale proposal layer that can generate object proposals of various scales. We integrate the R-DAD into several feature extractors, and prove the distinct performance improvement on PASCAL07/12 and MSCOCO18 compared to the recent convolutional detectors.

\end{abstract}

\section{Introduction}

Object detection is to find all the instances of one or more classes of objects given an image. In the recent years, the great progress of object detection have been also made  by combining the region proposal algorithms and CNNs. The most notable work is the R-CNN \cite{GirshickDDM_CVPR14} framework. They first generate object region proposals using the selective search \cite{UijlingsSGS_IJCV13}, extract CNN features \cite{KrizhevskySH_NIPS12} of the regions, and classify them with  class-specific  SVMs. Then, Fast RCNN \cite{Girshick15_ICCV15} improve the R-CNN speed  using feature sharing and RoI pooling.
The recent   detectors  \cite{RenHGS15_NIPS15,RedmonDGF_CVPR16,LiuAESRFB_ECCV16} integrate the external region proposal modules into a CNN for boosting the training and detection speed further. As a result, the detection accuracy can be also enhanced by joint  learning of region proposal and classification modules.

The modern convolutional detectors usually simplify feature extraction and object detection processes with a fixed input scale. But even with the robustness of the CNNs to the scale variance, the region proposal  accuracy is frequently degraded by the mismatches of produced proposals and object regions.  Also, the mismatch tends to be increased for the small object detection  \cite{LinDGHHB_Corr16}. To improve the proposal accuracy, multi-scale feature representation using feature pyramid is used for generating stronger synthetic feature maps. However, featurizing each level of an image pyramid increases the inference time significantly. In an attempt to reduce the detection complexity, \cite{LinDGHHB_Corr16,FuLRTB_Corr17} leverage the feature pyramid of CNNs.

In general, detections failures are  frequently caused for occluded objects.  In this case, the CNN feature discriminatbility for the occluded one can be reduced considerably since the some part details of the object are missing in the occluded regions. It implies that exploiting global appearance features for an entire object region could be insufficient to classify and localize objects accurately.

In this paper, we propose a novel region decomposition and assembly detector (R-DAD) to resolve the  limitations of the previous methods. The proposed method is based on (1) \textit{multi-scale-based region proposal} to improve region proposal accuracy of the region proposal network (RPN) and (2) \textit{multi-region-based appearance model} to describe the global and part appearances of an object jointly. 

In a multi-scale region proposal layer, we first generate region proposals using RPN and re-scale the proposals  with different scaling factors to cover the variability of the object size. We then select the region proposals suitable for training and testing in considerations of the ratios of object and non-object samples to handle the data imbalanced problem. The main benefits of our method is  that we can deal with the variability of the object size using the region scaling without expensive image or feature pyramids while maintaining the appropriate  number of region proposals using the region sampling. In addition, we can capture local and global context cues by rescaling the region proposals. To be more concrete, we can capture the  local details with smaller proposals than its original region  and  the global context between object and surround regions with larger proposals.

In order to improve the feature discriminability, we further perform multi-region based appearance learning by combining features of an entire object and its parts. The main idea behind this method is that a maximum response from each feature map is a strong visual cue to identify  objects. However, we also need to learn the semantic relations (\ie weights) between the entire and  decomposed regions for combining them adaptively. For instance, when the left part of an object is occluded, the weights should be adjusted to be used features of the right part more for object detection since the features of the less occluded part are more reliable. To this end, we propose a region assembly block (RAB) for ensembling multi-region features. Using the RABs, we first learn the relations between part feature maps and generate a combined feature map of part models by aggregating maximum responses of part features  stage-by-stage. We then produce strong high-level semantic features by combining global appearance and part appearance features, and use them for classification and localization.

To sum up, the main contributions of this paper can be summarized as follows: (i) proposition of the R-DAD architecture that can perform multi-scale-based region proposal and multi-region-based appearance learning through end-to-end training (ii) achievement of state-of-the-art results without employing  other performance improvement methods (\eg feature pyramid, multi-scale testing, data augmentation, model ensemble, etc.) for several detection benchmark challenge on PASCAL07 (mAP of 81.2\%), PASCAL12 (mAP of 82.0\%), and MSCOCO18 (\FN{mAP of 43.1\%}) (iii) extensive implementation of R-DADs with various feature extractors and thorough ablation study to prove the effectiveness and robustness of R-DAD. 
In this work, we first apply the proposed detection methods for the Faster RCNN \cite{RenHGS15_NIPS15}, but we believe that our methods can be applied for other covolutional detectors \cite{Girshick15_ICCV15,BellZBG_CVPR16,KongYCS_CVPR16,DaiLHS_CORR16} including RPN since these methods do not depend on a structure of a network.

\section{Related Works}

In spired of the recent advances in deep learning, a lot of progress has been made on object detection. In particular,  convolutional detectors  have become popular since  it allows effective feature extraction and end-to-end training from image pixels to object classification and localization. In particular, the remarkable improvement of deep networks for large scale object classification has also leaded to the improvement of detection methods. For feature extraction and object classification, the recent object detectors are therefore constructed based on the deep CNNs \cite{SimonyanZ14a,HeZRS_CVPR16} trained beforehand with large image datasets, and combined with region proposal and box regression networks for object localization.  Among several works, Faster-RCNN \cite{RenHGS15_NIPS15} achieve the noticeable performance improvement by integrating RPN and Fast RCNN  \cite{Girshick15_ICCV15}. In addition,  \cite{RedmonDGF_CVPR16,LiuAESRFB_ECCV16} develop the faster detectors by  predicting object class and locations from  feature maps directly without the region proposal stage.

For improving detection and segmentation, multiple feature maps extracted from different resolutions and regions have been exploited. 
\cite{ICCV15_gidaris} improve the feature discriminability and diversity by combining several region features. \cite{ZengOYLXWLZYWZW_CORR16} learn the relation and dependency between feature maps of different resolutions via message passing. \cite{LinDGHHB_Corr16}  connect convolved and deconvolved (\textit{or} up-sampled) feature mas from bottom-up and top-down pathways for multi-scale feature representation. HyperNet \cite{KongYCS_CVPR16} and ION \cite{BellZBG_CVPR16} concatenate different layer feature maps, and then predict the objects with the transformed maps having more contextual and semantic information. Basically, the previous works based on multiple feature maps focus on (1) multi-region representation to improving the feature discriminability and diversity  (2) multi-scale representation to detect the objects with small sizes without image pyramid. Although most previous detection methods with multiple features focus on only one of both issues, the proposed R-DAD can efficiently handle both issues together.

\begin{figure*}[!t]
\vspace{-0pt}
\begin{center}
\includegraphics[width=0.90\linewidth]{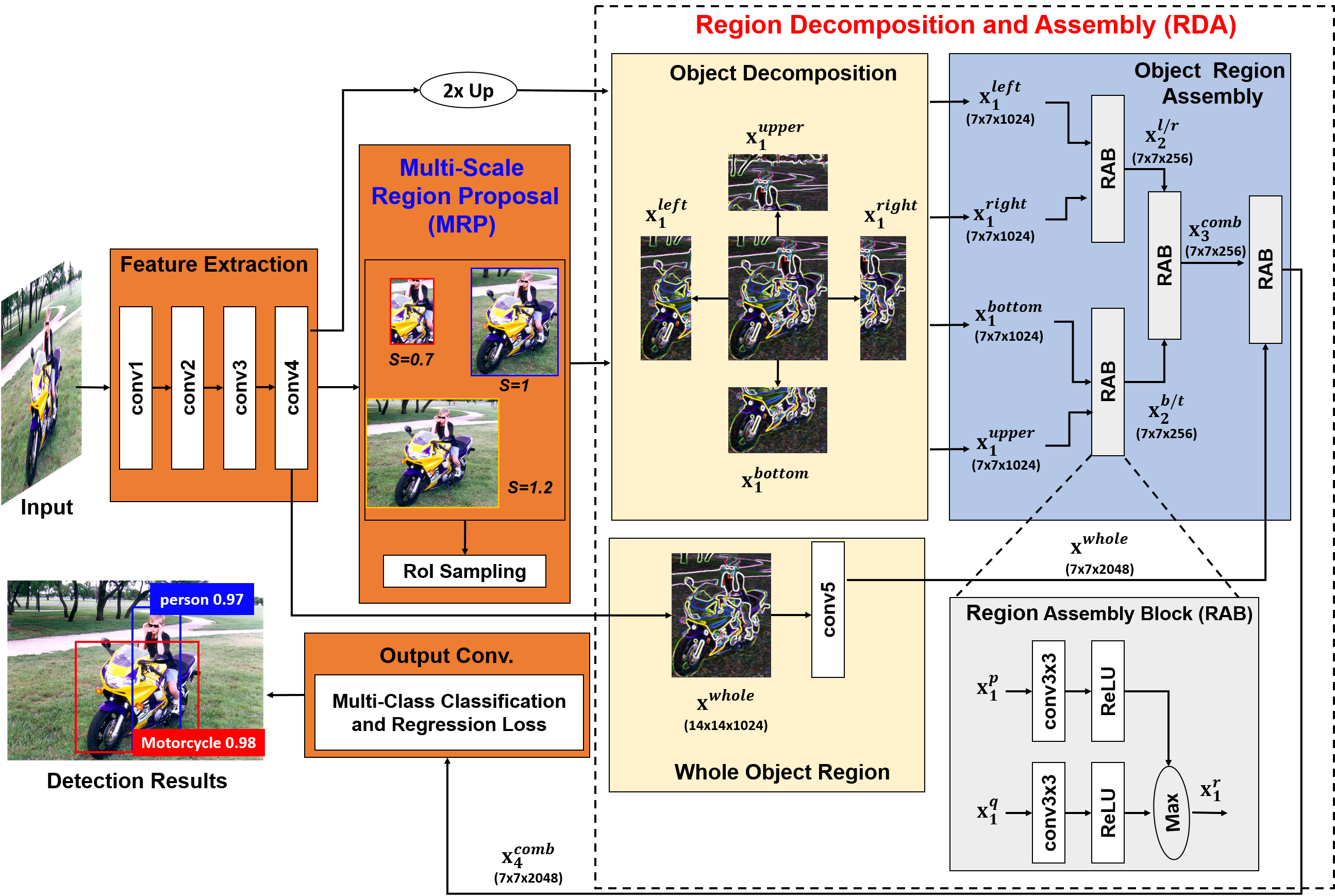}
\caption{{Proposed R-DAD architecture: In the MRP network, rescaled proposals are generated. \SH{For each rescaled proposal, we decompose it into several part} regions. We design a region assembly block (RAB) with 3x3 convolution filters, ReLU, and max units. In the RDA network, \SH{by using RABs we combine the strong responses of decomposed object parts  stage by stage}, and then learn the semantic relationship between the whole object and part-based features. }}
\vspace{-10pt}
\label{fig:MRD}
\end{center}
\end{figure*}

\section{Region Decomposition and Assembly Detector}

The architecture of our R-DAD is shown in Fig. \ref{fig:MRD}. It mainly consists of feature extraction, multi-scale-based region proposal (MRP) and object region  decomposition and assembly (RDA) phases. For extracting a generic CNN features, similar to other works, we use a classification network trained with \FN{ImageNet} \cite{ILSVRC15}. In our case, to prove the flexibility of our methods for the feature extractors, we implement different detectors by combining our methods with several feature extractors: ZF-Net \cite{ZeilerF_ECCV14}, VGG16/VGGM1024-Nets \cite{SimonyanZ14a}, Res-Net101/152 \cite{HeZRS_CVPR16}. In Table \ref{TABLE:Comp_FRCN}, we compare the detectors with different feature extractors. In Fig. \ref{fig:MRD}, we however design the architecture using ResNet as a base network.

In the MRP network, we generate region proposals (\ie bounding boxes) of different sizes. 
By using the RPN \cite{RenHGS15_NIPS15}, we first generate   proposal boxes. We then re-scale the generated proposals  with different scale factors for enhancing diversity of region proposals.  In the RoI sampling layer, we select appropriate boxes among them for training and inference in consideration of the data balance between foreground and background samples.

In addition, we learn the global (\ie an entire region) and part appearance (\ie decomposed regions) models  in the RDA network. The main process is that we decompose an entire object region into several small regions and extract  features of each region. We then merge the several part models while learning the strong semantic dependencies between decomposed parts. Finally, the combined feature maps between part and global appearance models are used for object regression and classification.

\subsection{Faster RCNN}
We briefly introduce Faster-RCNN \cite{RenHGS15_NIPS15} since we implement our R-DAD on this framework. The detection process of Faster-RCNN can be divided into two stages. In the first stage, an input image is resized to be fixed and is fed into a feature extractor (\ie pretrained classification network) such as VGG16 or ResNet. Then, the PRN uses mid-level features at some selected intermediate level (\eg ``conv4" and ``conv5" for VGG and ResNet) for generating class-agnostic box proposals and their confidence scores. 

In the second stage, features of box proposals are cropped by RoI pooling from the same intermediate feature maps used for box proposals. Since  feature extraction for each proposal is simplified by cropping the extracted feature maps previously without the additional propagation, the speed can be greatly improved. Then, the features \SH{for} box proposals are subsequently propagated in  other higher layers (\eg ``fc6" followed by ``fc7") \FN{to predict  a class and refine states} (\ie locations and sizes) for each candidate box.

\subsection{Multi-scale region proposal (MRP) network}
Each bounding box can be denoted as $\mathbf{d} = \left(x, y, w, h \right)$, where $x$, $y$, $w$ and $h$ are the center positions, width and height. Given a region proposal $\mathbf{d}$, the rescaled box is $\mathbf{d}^{s} = \left(x, y, w \cdot s,   h \cdot s \right)$ with a scaling factor $s$ ($\geq$ 0). By applying different $s$ \SH{to} the original box, we can generate  $\mathbf{d}^{s}$ with different sizes. Figure \ref{fig:App} shows the rescaled boxes with different $s$, where the original box \SH{$\mathbf{d}$} has $s=1$.  In our implementation, we use different $s=[0.5, 0.7, 1, 1.2, 1.5]$. 

Even though boxes with different scales can be generated by the RPN in the Faster RCNN, we can further increase diversity of region proposals by using the multi-scale detection. By using larger $s$, we can capture contextual information (\eg background or an interacting object) around the object. On the other hand,  by using smaller $s$ we can investigate local details in higher resolution and it can be useful for identifying the object under occlusion where the complete object details are unavailable. The effects of the multi-scale proposals with different $s$ are shown in Table \ref{TABLE:Ablation_mmd}.

Since huge number of proposals (63$\times$38$\times$9$\times$5) are generated for the feature maps of size $63 \times 38$ at the ``conv4" layer when using 9 anchors and 5 scale factors,  exploiting all the proposals for training a network is impractical. Thus, we maintain the appropriate number of proposals (\eg 256) by removing the proposals with low confidence and low overlap ratios over ground truth. We then make a ratio of object and non-object samples in a mini-batch to be equal and use the mini-batch for fine-tuning a detector shown in Fig \ref{fig:MRD}.

\subsection{Region decomposition and assembly (RDA) network}
In general, strong responses of features is one of the most important cues to recognize objects. For each proposal from the MRP network, we therefore infer strong cues by combining features of multiple regions stage-by-stage as shown in Fig. \ref{fig:MRD}. To this end, we need to learn the weights which can \SH{represent} semantic relations between the features of different parts, and using the weights we control the amount of features to be propagated in the next layer.

\begin{figure}[!tbp]
\begin{center}
\vspace{-0pt}
\includegraphics[width=0.99\linewidth]{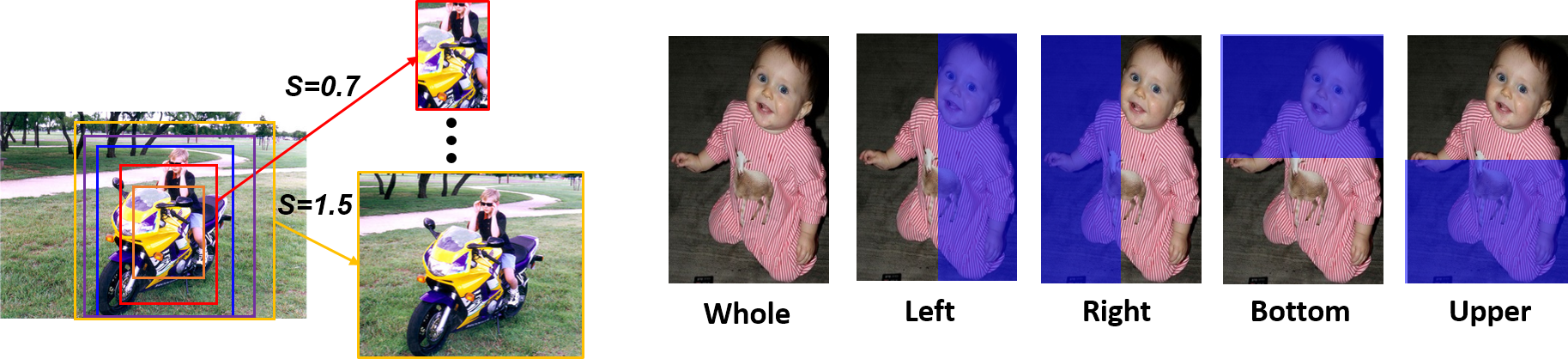}
\caption{{(Left) Rescaled \FN{proposals} by the MRP. (Right) Several decomposed regions for a whole object region.}}
\vspace{-10pt}
\label{fig:App}
\end{center}
\end{figure}

A region proposal   from RPN is assumed usually to cover the whole region of an object. We generate smaller decomposed regions by diving \SH{$\mathbf{d}$} into several part regions. We make the region cover different object parts as shown in Fig. \ref{fig:App}. 

From the feature map used as the input of the MRP network, we first extract the warped features $\mathbf{x}_{l}$ of size $h_{roi} \times w_{roi}$ for the whole object region using RoI pooling \cite{Girshick15_ICCV15}, where $h_{roi}$ and $w_{roi}$ are the height and the width of the map (for ResNet, $h_{roi}=14$ and $w_{roi}=14$).  Before extracting features of each part, we first upsample the spatial resolution of the feature map by a factor of 2 using bilinear interpolation. We found that these finer resolution feature maps improve the detection rate since the object details can be captured more accurately as shown in Table \ref{TABLE:Ablation_mmd}. Using RoI pooling, we also extract warped feature maps of size $\lceil h_{roi}/2 \rceil \times \lceil w_{roi}/2 \rceil$, and denote them as $\mathbf{x}^{p}_{l}, p \in \{\text{left, right, bottom, upper} \}$.

In the forward propagation, we  \SH{convolve} part features $\mathbf{x}_{i,l-1}^{p}$ at layer $l-1$ of size $h_{l-1}^{p} \times w_{l-1}^{p}$ with different kernels $\mathbf{w}_{ij}^{l}$ of size $m_{l} \times m_{l}$, and then \SH{pass the convolved features} a nonlinear activation function  $f(\cdot)$ to obtain an updated feature map $\mathbf{x}_{j,l}^{p}$ of size $(h_{l-1}^{p} - m_{l} +1) \times (w_{l-1}^{p} - m_{l} +1)$ as 
\vspace{-0pt} 
\begin{equation}\label{eq:update} 
\begin{array}{l} 
   \mathbf{x}_{j,l}^{p} = {f \left(\sum_{i=1}^{k_{l}}\mathbf{x}_{i,l-1}^{p} \ast \mathbf{w}_{ij}^{l} + \FN{b_{j}^{l}} \right)},~$l= 2, 3, 4$ 
\end{array}\vspace{-0pt} 
\end{equation}
where $p$ represent each part (left, right, bottom, upper) or combined parts (left-right(l/r), bottom-upper (b/u) and comb) \SH{as in Fig. \ref{fig:MRD}}. \FN{$b_{j}^{l}$} is a bias factor, $k_{l}$ is the number of kernels. $\ast$ means convolution operation. We use the element-wise ReLU function as $f(\cdot)$ for scaling the linear input.  

Then,  the bi-directional outputs \SH{$\mathbf{x}_{l}^{p}$ and $ \mathbf{x}_{l}^{q}$} Eq. (\ref{eq:update}) of different regions  are merged to produce the combined feature  $\mathbf{x}_{l}^{r}$  \SH{by using} an element-wise max unit over  each channel as 
\vspace{-0pt} 
\begin{equation}\label{eq:Fconv} 
\begin{array}{l} 
 \mathbf{x}_{l}^{r} =  \mathlarger{\text{max} \left(\mathbf{x}_{l}^{p}, \mathbf{x}_{l}^{q} \right)} 
\end{array}\vspace{-0pt} 
\end{equation}
$p$, $q$ and $r$ also represent each part or a combined part as shown in Fig. \ref{fig:App}. The element-wise max unit is used to merge  information between $\mathbf{x}_{l}^{p}$ and $\mathbf{x}_{l}^{q}$ and produce $\mathbf{x}_{l}^{r}$ with the same size. As a result, the bottom-up feature maps are refined state-by-stage by comparing features of different regions and strong semantics features remained only. The holistic feature of the object is propagated through several layers (in ``conv5 block" for ResNet) of the base network, and the features $\mathbf{x}^{whole}$ for the whole object appearance at the last layer   is also compared with the combined feature $\mathbf{x}_{3}^{comb}$ of part models, and then the refined features $\mathbf{x}_{4}^{comb}$ are connected with the object classification and box regression layers with $cls+1$ neurons and $4(cls+1)$ neurons, where $cls$ is the number of object classes and the one is added due to the background class.

\begin{figure}[!tbp]
\vspace{-0pt}
\begin{center}
\includegraphics[width=0.99\linewidth]{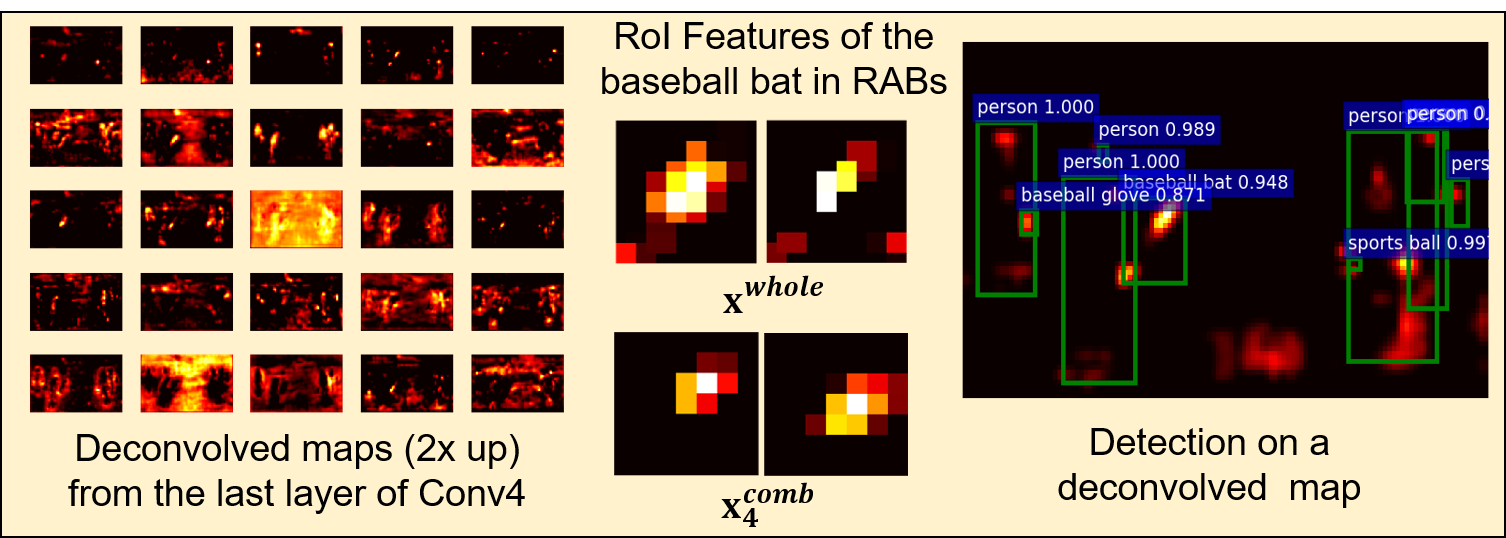}
\vspace{-10pt}
\caption{{Intermediate semantic features and detection results generated by  our R-DAD. }}
\vspace{-10pt}
\label{fig:Feature}
\end{center}
\end{figure}

Figure \ref{fig:Feature} shows the  semantic features at several layers of the learned R-DAD. Some strong feature responses within objects are extracted by our R-DAD.

\subsection{R-DAD Training}
\label{sub-sec:Training}
For training our R-DAD, \SH{we exploit the pre-trained and shared parameters  of  a feature extractor (\eg ``conv1-5'' of ResNet) from the \FN{ImageNet}  dataset as initial parameters of R-DAD.} We then fine-tune parameters  of higher layers (``conv3-5'') of the R-DAD  while keeping parameters of lower layers (\eg ``conv1-2"). We freeze the parameters for batch normalization which was learned during \FN{ImageNet} pre-training.  The parameters of the MRP and RDA networks are initialized  with the Gaussian distribution.

  For each  box \SH{$\mathbf{d}$}, we find the best matched ground truth box \SH{$\mathbf{d}^{\ast}$} by evaluating IoU. If a box \SH{$\mathbf{d}$} has an IoU than 0.5 with any   \SH{$\mathbf{d}^{\ast}$}, we assign positive label $o^{\ast}\ \in \{1...cls\}$, and a vector representing the 4 parameterized coordinates of  \SH{$\mathbf{d}^{\ast}$}. We assign a negative label (0) to \SH{$\mathbf{d}$} that has an IoU between 0.1 and 0.5.  From the output layers of the R-DAD, 4 parameterized coordinates and the class label $\hat{o}$  are predicted to each box \SH{$\mathbf{d}$}. The adjusted box  \SH{$\hat{\mathbf{d}}$} is generated by applying  the predicted regression parameters. For box regression, we use the following parameterization    \cite{GirshickDDM_CVPR14}.
\vspace{-0pt}
\begin{equation}\label{eq:Loss} 
\begin{array}{ll} 
t_{x} = \left(\hat{x} - {x} \right)/w, & t_{y} = \left(\hat{y} - {y} \right)/h, \\
t_{w} = \text{log} (\hat{w}/ {w} ), & t_{h} = \text{log}(\hat{h} /{h} ), \\
\end{array}
\end{equation}
where $\hat{x}$ and $x$ are for the predicted and anchor box, respectively (likewise for $y$, $w$, $h$). 
Similarly, $\mathbf{t}^{\ast}=[t_x^{\ast}, t_y^{\ast}, t_w^{\ast}, t_h^{\ast}]$ is evaluated with the predicted box and ground truth boxes.
We then train the R-DAD by  minimizing the classification and regression losses Eq.(\ref{eq:Loss}).
\begin{equation}\label{eq:Loss} 
\begin{array}{lc} 
L\left(\mathbf{o}, \mathbf{o}^{\ast}, \mathbf{t}, \mathbf{t}^{\ast}  \right) = L_{cls}(\mathbf{o}, \mathbf{o}^{\ast}) + \lambda \left[o \geq 1 \right]L_{reg}(\mathbf{t}, \mathbf{t}^{\ast})
\end{array}
\end{equation}
\begin{equation}\label{eq:Loss_cls} 
\begin{array}{lc} 
 L_{cls}(\mathbf{o}, \mathbf{o}^{\ast}) = - {\sum}_{u} \delta(u, o^{\ast}) log(p_{u}), 
\end{array}
\end{equation}
\begin{equation}\label{eq:Loss_reg} 
\begin{array}{lc} 
 L_{loc}(\mathbf{t}, \mathbf{t}^{\ast}) = - {\sum}_{v \in \{x,y,w,h\}} \text{smooth}_{L_{1}}\left(t_{v}, t_{v}^{\ast} \right),  
\end{array}
\end{equation}
\begin{equation}\label{eq:smooth} 
\begin{array}{lc} 
 \text{smooth}_{L_{1}} (z) = \begin{cases}
0.5z^{2} & \text{ if } \left|z \right| \leq 1\\ 
 \left|z \right| - 0.5& \text{ otherwise } 
\end{cases} 
\end{array}
\end{equation}
where $p_{u}$ is the predicted classification probability for class $u$. $\delta(u, o^{\ast}) =1 $ if $u = o^{\ast}$ and  $\delta(u, o^{\ast}) =0$ otherwise. Using the SGD with momentum of 0.9, we train the parameters.  $\lambda = 1$ in our implementation.

\vspace{-0pt}
\subsection{Discussion of R-DAD}
\vspace{-0pt}

There are several benefits of the proposed  object region assembly method. Since we extract maximum responses across spatial locations between feature  maps of object parts, we can  improve the spatial invariance more to feature position without a deep hierarchy than using max pooling which supports a small region (\eg $2 \times 2$). Since our RAB is similar to the maxout unit  except for using ReLU, the RAB can be used as a universal approximator which can approximate arbitrary continuous function \cite{GoodfellowWMCB_ICML13}. This indicates that a wide variety of object feature configurations can be represented  by combining our RABs hierarchically. In addition, the variety of region proposals generated by the MRP network can improve the robustness further to  feature variation occurred by the \SH{spatial} configuration change between objects.

\begin{table}[!tbp]
\vspace{-0pt}
\caption{{Ablation study: effects of the proposed multi-scale region proposal and object region decomposition/assembly methods.}}  
\vspace{-0pt}
{\scriptsize
\begin{center}
\renewcommand{\tabcolsep}{0.7mm}
\begin{tabular}{c|ccccccc}
\hline \hline
\textbf{Method} & \multicolumn{7}{c} {\textbf{Combination}}\\
\hline 
Multi-scale region proposal &  & \multirow{2}{*}{$\surd$} & & & \multirow{2}{*}{$\surd$} & & \\ 
$s=[0.7, 1.0, 1.5]$ &  & & & &  & & \\ \hline
Multi-scale region proposal &  & &  \multirow{2}{*}{$\surd$} & &  & \multirow{2}{*}{$\surd$} & \multirow{2}{*}{$\surd$} \\ 
$s=[0.5, 0.7, 1.0, 1.2, 1.5]$ &  & & & & & &  \\  \hline
Decomposition/assembly & &  & & $\surd$& $\surd$ &  $\surd$ & $\surd$ \\  \hline
Up-sampling & &  & & &  &   &$\surd$ \\  \hline
\textbf{Mean AP}& 68.90 & 70.0 & 70.30 & 71.95 & 72.65 &73.90 &\underline{\textbf{74.90}} \\
\hline \hline 
\end{tabular}
\end{center}}
\label{TABLE:Ablation_mmd}
\vspace{-10pt}
\end{table}

\section{Implementation}
\label{sec:Imp}
We apply our R-DAD for \SH{various} feature extractors to show \SH{its flexibility to the base network}. In general, a feature extractor  affects the  accuracy and speed of a detectors. In this work, we use five different feature extractors and combine each extractor with our R-DAD to compare \SH{each other} as shown in Table \ref{TABLE:Comp_FRCN} and \ref{TABLE:speed}. We implement all the detectors using  the Caffe  on a PC with a single TITAN Xp GPU  without parallel and distributed training. 

\subsection{ZF and VGG networks}
We use the fast version  of ZF \cite{ZeilerF_ECCV14} with 5 shareable convolutional  and 3 fully-connected layers. We also use the VGG16 \cite{SimonyanZ14a} with 13 shareable convolutional  and 3 fully connected layers. Moreover, we exploit the VGGM1024 (variant of VGG16) with the same depth of AlexNet \cite{KrizhevskySH_NIPS12}. All these models pre-trained with the ILSVRC classification dataset are given by \cite{RenHGS15_NIPS15}.

To generate region proposals, we feed the feature maps of the last shared convolutional layer (``conv5" for ZF and VGGM1024, and ``conv5\_3" for VGG-16) to the MRP network. Given a set of region proposals, we also feed the shared maps of the last layer to the RDA network for learning high-level semantic features by combining the features of decomposed regions. We use \SH{$\mathbf{x}_{4}^{comb}$} produced by the RDA network as inputs of regression and classification layers. We fine-tune all layers of ZF and VGG1024, and conv3\_1 and up for VGG16 to compare our R-DAD with Faster RCNN \cite{RenHGS15_NIPS15}. The sizes of a mini-batch  used for training MRP and RAD networks are set to 256 and 64, respectively. 

\subsection{Residual networks}
We use the  ResNets \cite{HeZRS_CVPR16} with different depths by stacking different number of residual blocks. For the ResNets50/101/152 (Res50/101/152), the layers from ``conv1" to ``conv4" blocks are shared in the Faster RCNN.  In a similar manner, we use the features from the last layer of the ``conv4" block as inputs of the MRP and RDA networks. We fine-tune the layers of MRP and RDA networks including layers of ``conv3-5''  while freezing layers of ``conv1-2'' layers. We also use the same mini-batch sizes (256/64) when training MRP and RDA networks per iteration.

\section{Experimental results}
We train and evaluate our R-DAD on standard detection benchmark datasets: PASCAL VOC07/12 \cite{Everingham15} and MSCOCO18 \cite{LinMBHPRDZ_corr14}  datasets. We first evaluate the proposed methods with ablation experiments. We then compare the R-DAD with modern detectors on the datasets.

\textbf{Evaluation measure:}
We use average precision (AP) per class which is a standard metric for object detection. It is evaluated by computing the area under the precision-recall curve. We also compute mean average precision (mAP) by averaging the APs over all object classes. When evaluating AP and mAP on PASCAL and COCO, we use the public available codes \cite{Girshick15_ICCV15,LinMBHPRDZ_corr14} or evaluation servers for those competition.

\textbf{Learning strategy:}
We use different learning rates for each  evaluation. 
We use a learning rate $\mu=1\text{e}^{-3}$ for 50k iterations, and $\mu=1\text{e}^{-4}$ for the next 20k iterations on VOC07 evaluation. For VOC12 evaluation,  we train a detector with $\mu=1\text{e}^{-3}$ for 70k iterations, and continue it for 50k iterations with $\mu=1\text{e}^{-4}$. For MSCOCO  evaluation, we use $\mu=1\text{e}^{-4}$ and $\mu=1\text{e}^{-5}$ for the first 700k  and  the next 500k iterations.

\begin{table}[tbp]
\vspace{-0pt}
\caption{Ablation study: the detection comparison of different region assembly blocks.}  
\vspace{-5pt}
{\scriptsize
\begin{center}
\renewcommand{\tabcolsep}{0.7mm}
\begin{tabular}{c|c|c|c}
\hline \hline
\textbf{Stage 1} & \textbf{Stage 2}  & \textbf{Stage 3}  & \textbf{Mean AP} \\ \hline
Sum& Sum  & Sum  & 69.61 \\
Sum&Max&Max	&69.34\\
Sum&Sum&Max	&69.82\\
Max&Max&Sum	&69.08\\
Max&Max&Sum $[{c}_1=1,c_2= \gamma]$	&68.80\\
Max&Max&Sum $[{c}_1=\gamma,c_2= 1]$	&69.30\\
Max&Max&Concatenation	&71.95\\
Max(dil. $d=4$)&Max&Max	&70.87\\
Max (dil. $d=2$)&Max(dil. $d=2$)&Max	&70.55\\
Max&Max(dil. $d=4$)&Max	&70.64\\
Max($m=5$)&Max ($m=5$)&Max	&75.10\\
\underline{\textbf{Max ($m=3$)}} &\underline{\textbf{Max ($m=3$)}}&\underline{\textbf{Max ($m=3$)}}	&\underline{\textbf{74.90}}\\
\hline 
\hline  
\end{tabular}
\end{center}}
\vspace{-15pt}
\label{TABLE:ILSVRC_Ab}
\end{table}

\subsection{Ablation experiments}
To show the effectiveness of the methods used for MRP and RDA networks, we perform several ablation studies. For this evaluation, we train detectors with  VOC07+(trainval) set and test them on the VOC07 test set.

\textbf{Effects of proposed methods:}
In Table. \ref{TABLE:Ablation_mmd}, we first show mAPs of a baseline detector without the multi-scale and region decomposition/assembly methods (\ie Faster RCNN) and the detector by using the proposed methods. Compared to the mAP of the baseline, we achieve the better rates when applying our methods. We also compare mAP of detectors with different number of scaling factors.  By using five scales, mAP is improved slightly. In particular, using decomposition/assembly method can improve mAP to \FN{$ 3.05\%$}.   Using up-sampling improves mAP to $1\%$. This indicates that  part features extracted in finer resolution yield the better detection.  As a result, combining all proposed methods with the baseline enhances the mAP to 6\%. 

\textbf{Structure of the RDA network:}
To determine the best structure of the RDA network, we evaluate mAP  by changing its components  as shown in Table \ref{TABLE:ILSVRC_Ab}. We first compare feature aggregation methods. As  in Fig.  \ref{fig:MRD}, we  combine the bi-directional outputs of different regions at each stage using  a max unit. We change this unit one-by-one  with  sum or concatenation units. When summing both outputs at the stage 3, we try to merge outputs with different coefficients.  Sum \FN{$[{c}_1=\gamma,c_2= 1]$} means that $\mathbf{x}^{whole}$ and $\mathbf{x}_{3}^{comb}$ are summed with $\gamma$ and \FN{1} weights. This is a similar concept to the identity mapping \cite{HeZRS_CVPR16}. The scale parameter $\gamma$ is learned during training. 
However, we found that summing  feature maps  or using identity mapping show the minor improvement. In addition, concatenating features improves the mAP, but it also increases the memory usage and complexity  of convolution at the next layer. From this comparison, we verify that  merging the features using the max units for all the stages provides the best mAP while  the computational complexity.   This evaluation  supports that our main idea of that maximum responses of features  are strong visual cues for detecting objects.

Moreover, to determine the effective receptive field size, we change the size of convolution kernels with $m=5$ at the stage 1 and 2 in the RDA network. Moreover, we also \SH{try} $d$-dilated convolution filters to expand the receptive field more. However, exploiting the dilated convolutions and 5x5 convolution filters  does not increase the mAP significantly. It indicates that  we can cover the  receptive \FN{fields} of each part and \SH{combined regions sufficiently} with the 3x3 filters.

\subsection{Comparison with Faster-RCNN}
\label{sub-sec:PASCAL07}
\textbf{Accuracy:}
To compare the Faster-RCNN (FRCN), we  train both detectors with the VOC07trainval (\textbf{VOC07}, 5011 images) and VOC12trainval  sets (\textbf{VOC07++12}, 11540 images). We then test them on  the VOC07 test set (4952 images). 
For more comparison, we implement both detectors with various feature extractors. The details of the implementation are mentioned in previous section. Table \ref{TABLE:Comp_FRCN} shows the comparison results of both detectors. All the R-DADs show the better mAPs than those of Faster RCNNs.  We improve mAP about 3$\sim$ 5\% using R-DAD. We also confirm that using feature extractors with higher classification accuracies leads to better detection rate.

\begin{table}[tbp] \vspace{-5pt}
\caption{{Comparison between the R-DAD and Faster-RCNN by  using different feature extractors on the VOC07 test set.}} 
 \vspace{-0pt}
{\scriptsize
\begin{center}
\renewcommand{\tabcolsep}{0.5mm}
\begin{tabular}{c|c|c||c|c|ccc}
\hline \hline
 \textbf{Train set} & \textbf{Detector} &   \textbf{mAP} &  \textbf{Train set} & \textbf{Detector} &   \textbf{mAP} \\
\hline \hline
&FRCN/ZF& 60.8 & &FRCN/ZF&66.0 \\ 
&{\textbf{R-DAD/ZF}} & \underline{\textbf{63.7}} & &\textbf{R-DAD/ZF} & \underline{\textbf{68.2}} \\  \cline{2-3}  \cline{5-6}
PASCAL&FRCN/VGGM1024& 	61.0 & PASCAL&FRCN/VGGM1024& 	65.0\\
VOC&{\textbf{R-DAD/VGGM1024}} &\underline{\textbf{65.0}} & VOC&\textbf{R-DAD/VGGM1024} &\underline{\textbf{69.1}} \\  \cline{2-3}  \cline{5-6}
\textbf{07}&FRCN/VGG16&69.9 & \textbf{07++12}&FRCN/VGG16& 73.2\\
&\textbf{R-DAD/VGG16} &\underline{\textbf{73.9}} & & \textbf{R-DAD/VGG16} &\underline{\textbf{78.2}} \\  \cline{2-3}  \cline{5-6}
&FRCN/Res101&74.9 & &FRCN/Res101& 76.6\\
&\textbf{R-DAD/Res101} &\underline{\textbf{77.6}} & & \textbf{R-DAD/Res101} &\underline{\textbf{81.2}} \\  \cline{2-3}  \cline{5-6}
\hline \hline
\end{tabular}
\vspace{-15pt}
\end{center}}
\label{TABLE:Comp_FRCN}
\end{table}

\begin{table*}[!tbp]
\vspace{-0pt}
\centering
\caption{The speed of the Faster R-CNN (FRCN) and  R-DAD (input size: 600 $\times$1000).}
\vspace{-0pt}
\label{TABLE:speed}
\resizebox{\textwidth}{!}{%
\begin{tabular}{|c|c|c|c|c|c|c|c|c|c|c|c|c|}
\hline
\textbf{Base Network}    & \multicolumn{2}{c|}{\textbf{ZF}} & \multicolumn{2}{c|}{\textbf{VGGM1024}} & \multicolumn{2}{c|}{\textbf{VGG16}} & \multicolumn{3}{c|}{\textbf{Res101}} & \multicolumn{3}{c|}{\textbf{Res152}} \\ \hline
\textbf{Detector}        & FRCN            & R-DAD          & FRCN               & R-DAD             & FRCN             & R-DAD            & FRCN        & R-DAD  & R-DAD($m=5$)   & FRCN         & R-DAD   & R-DAD($m=5$) \\ \hline
\textbf{Time(sec/frame)} & 0.041           & 0.048          & 0.046              & 0.054             & 0.15             & 0.177            & 0.208    & 0.245 & 0.53               & 0.301    & 0.385  & 0.574            \\ \hline
\end{tabular}}
\end{table*}

\textbf{Speed:}
In Table \ref{TABLE:speed}, we have compared the detection speed of both detectors. Since the speed depends on size of the base network, we evaluate them by using various base networks. We also fix the number of region proposals to 300 as done in \cite{Girshick15_ICCV15}. The speed of our R-DAD  is comparable with it of the FRCN. Indeed, to reduce the detection complexity while maintaining the accuracy, we  design the R-DAD structure in consideration of several important factors. We found that the spatial sizes of RoI feature maps ($h_{roi}$ and $w_{roi}$)  and convolution filters (m) can affect the speed significantly. When using  $h_{roi}=14$, $w_{roi}=14$ and $m=5$ in RABs, R-DAD gets 1.5x $\sim$ 2.1x slower but enhanced only about \FN{0.2\%} as in Table \ref{TABLE:ILSVRC_Ab}.  {Therefore, we confirm that adding MRP and RDA networks to the Faster RCNN does not increase the complexity significantly.} 

\section{Detection Benchmark Challenges}
In this evaluation, our R-DAD performance is evaluated from PASCAL VOC and MSCOCO  servers. We also post our detection scores to the leaderboard of each challenge.

\textbf{PASCAL VOC 2012 challenge:}
We  evaluate our R-DAD on the PASCAL VOC 2012 challenge.
For training our R-DAD, we use VOC07++12 only and test it on VOC2012test (10911 images). Note that we do not use extra datasets such as the COCO dataset for improving mAP  as done in many top ranked teams.

Table \ref{TABLE:Comp_PASCAL12} shows the results. As shown, we achieve the best mAP among state-of-the-art convolutional detectors. In addition, our detector shows the higher mAP  \FN{by} using the Res152 model. Compared to the Faster RCNN and MR-CNN \cite{ICCV15_gidaris} using multi-region approach, we improve the mAP to 11.6\% and 8.1\%.

\begin{table*}[!t] \vspace{-10pt}
\caption{Performance comparison with other detectors in PASCAL VOC 2012 challenge. The more results  can be found in  \href{http://host.robots.ox.ac.uk:8080/leaderboard/displaylb.php?challengeid=11&compid=4}{the PASCAL VOC 2012 website}.}
 \vspace{-5pt}
{\footnotesize
\begin{center}
\renewcommand{\tabcolsep}{0.7mm}
\resizebox{\textwidth}{!}{%
\begin{tabular}{c|c|c|ccccccccccccccccccccccc}
\hline \hline
 \textbf{Train set} & \textbf{Detector} &   \textbf{mAP} & \textbf{aero} & \textbf{bike} &\textbf{bird}  &\textbf{boat} & \textbf{bottle} & \textbf{bus} &\textbf{car}  &\textbf{cat}  &\textbf{chair} &\textbf{cow} &\textbf{table} &\textbf{dog} &\textbf{horse} &\textbf{mbike} &\textbf{person} &\textbf{plant} &\textbf{sheep} &\textbf{sofa} &\textbf{train} &\textbf{tv}  \\
\hline \hline
&Fast \cite{Girshick15_ICCV15} & 68.4 &82.3 &78.4 &70.8 &52.3 &38.7 &77.8 &71.6 &89.3 &44.2 &73.0 &55.0 &87.5 &80.5 &80.8 &72.0 &35.1 &68.3 &65.7 &80.4 &64.2\\
&Faster  \cite{RenHGS15_NIPS15} & 70.4 &84.9 &79.8 &74.3 &53.9 &49.8 &77.5 &75.9 &88.5 &45.6 &77.1 &55.3 &86.9 &81.7 &80.9 &79.6 &40.1 &72.6 &60.9 &81.2 &61.5\\
&SSD300 \cite{LiuAESRFB_ECCV16} &72.4 &85.6 &80.1 &70.5 &57.6 &46.2 &79.4 &76.1 &89.2 &53.0 &77.0 &60.8 &87.0 &83.1 &82.3 &79.4 &45.9 &75.9 &69.5 &81.9 &67.5 \\
PASCAL&SSD512 \cite{LiuAESRFB_ECCV16} &74.9 &87.4 &82.3 &75.8 &59.0 &52.6 &81.7 &81.5 &90.0 &55.4 &79.0 &59.8 &88.4 &84.3 &84.7 &83.3 &50.2 &78.0 &66.3 &86.3 &72.0\\
VOC&YOLOv2 \cite{RedmonF_CVPR17} &73.4 &86.3 &82.0 &74.8 &59.2 &51.8 &79.8 &76.5 &90.6 &52.1 &78.2 &58.5 &89.3 &82.5 &83.4 &81.3 &49.1 &77.2 &62.4 &83.8 &68.7\\
\textbf{07++12}&MR-CNN \cite{ICCV15_gidaris}&73.9 &85.5 &82.9 &76.6 &57.8 &62.7 &79.4 &77.2 &86.6 &55.0 &79.1 &62.2 &87.0 &83.4 &84.7 &78.9 &45.3 &73.4 &65.8 &80.3 &74.0 \\
&HyperNet \cite{KongYCS_CVPR16}& 71.4 &84.2 &78.5 &73.6 &55.6 &53.7 &78.7 &79.8 &87.7 &49.6 &74.9 &52.1 &86.0 &81.7 &83.3 &81.8 &48.6 &73.5 &59.4 &79.9 &65.7 \\
&ION \cite{BellZBG_CVPR16}& 76.4 &87.5 &84.7 &76.8 &63.8 &58.3 &82.6 &79.0 &90.9 &57.8 &82.0 &64.7 &88.9 &86.5 &84.7 &82.3 &51.4 &78.2 &69.2 &85.2 &73.5 \\
&\textbf{R-DAD/Res101} & \underline{\textbf{80.2}} &90.0& 86.6& 81.3& 71.2& 66.0& 83.4& 83.7& 94.5& 63.2& 84.0& 64.2& 92.8& 90.1& 88.6& 87.3& 62.2& 82.8& 70.9& 88.8& 72.2 \\
&\textbf{R-DAD/Res152} & \underline{\textbf{82.0}}& 90.2& 88.1& 85.3& 73.3& 71.4& 84.5& 87.4& 94.6& 65.1& 86.8& 64.0& 94.1& 89.7& 89.2& 89.3& 64.5& 83.5& 72.2& 89.5& 77.6 \\
\hline \hline
\end{tabular}}
\vspace{-15pt}
\end{center}}
\label{TABLE:Comp_PASCAL12}
\end{table*}

\textbf{MS Common Objects in Context (COCO) 2018 challenge:}
We  participate  in the MSCOCO  challenge. This challenge is detection for 80 object categories. We use COCO-style evaluation metrics: mAP averaged for IoU $\in [0.5:0.05:0.95]$, average precision/recall on small (\textbf{S}), medium  (\textbf{M}) and large (\textbf{L}) objects, and average recall on the number of detections (\# Dets).
We train our R-DAD with the union of train and validation images (123k images). We then test it on the test-dev set (20k images).  For enhancing  detection for the small objects, we use 12   anchors consisting of 4 scales (64, 128, 256, 512)  and 3 aspect ratios (1:1, 1:2, 2:1).

Table \ref{TABLE:COCO} compares the performance of detectors based on a single network. We divide detectors with single-stage-based and two-stage-based  detectors depending on region proposal approach. Note that our R-DAD with ResNet-50 is superior to other detectors. The performance of R-DAD is further improved to 40.4\% by using \SH{ResNet-101} with higher accuracy. 

Compared to the scores of this challenge winners of Faster R-CNN+++ (2015) and G-RMI (2016), our detectors produce the \SH{better results}. Remarkably,  we achieve the best scores without bell and whistles  (\eg multi-scale testing, hard example mining \cite{ShrivastavaGG_CVPR16}, feature pyramid \cite{LinDGHHB_Corr16}, model ensemble, etc).  By applying multi-scale testing for R-DADs with ResNet50 and ResNet101, we can improve mAP to 41.8\% and 44.9\%, respectively. As shown in this challenge leaderboard, our R-DAD is ranked on the high place.

\begin{figure}[!tbp]
\vspace{-5pt}
\begin{center}
\includegraphics[width=0.99\linewidth]{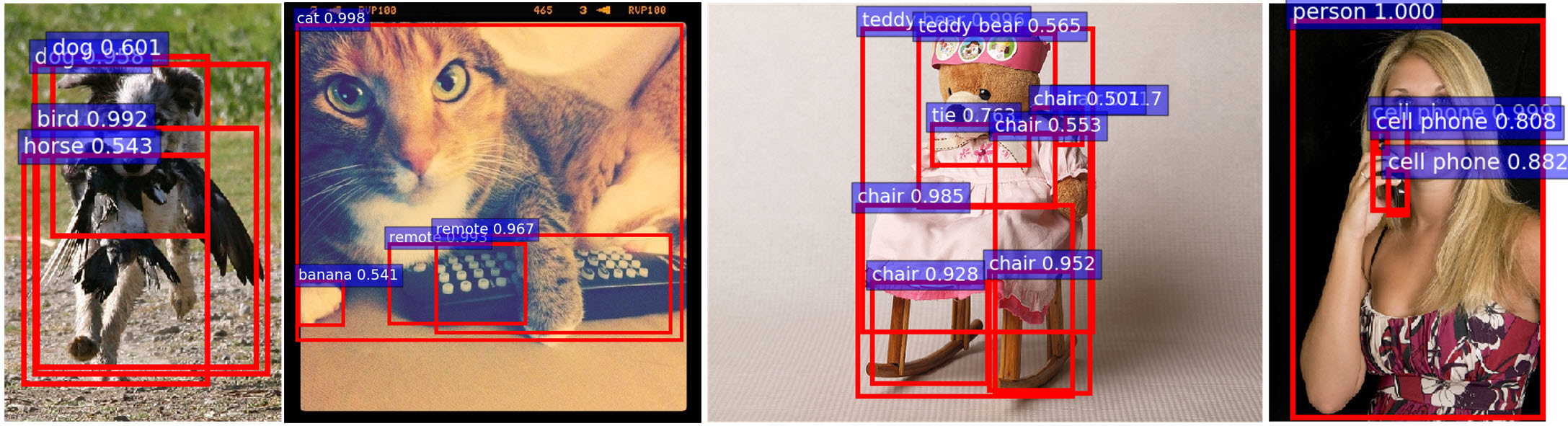}\\
\includegraphics[width=0.99\linewidth]{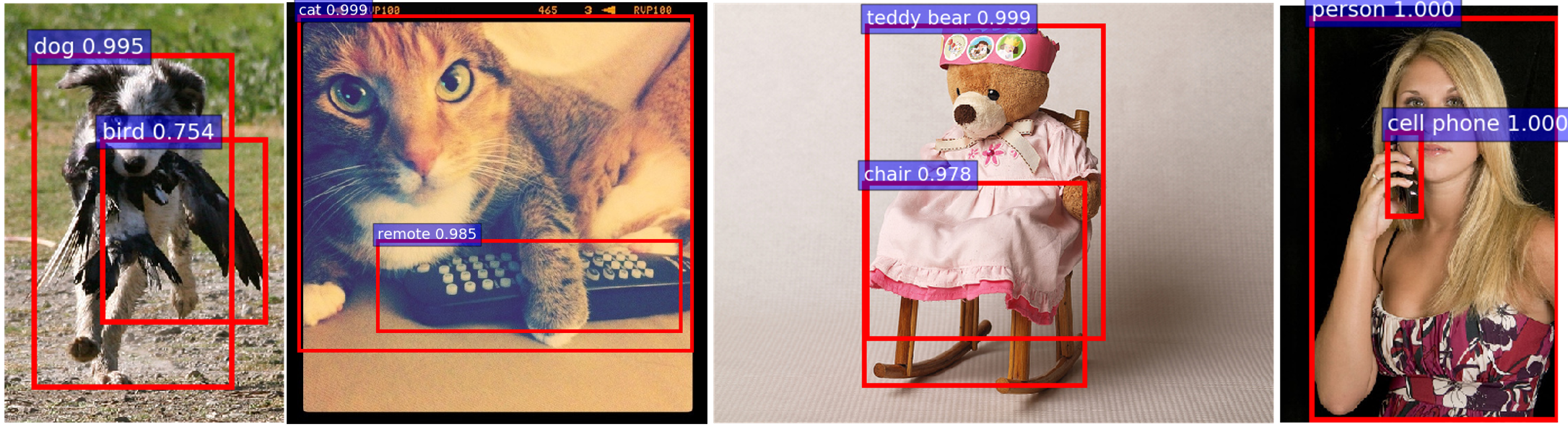}\\
\vspace{-10pt}
\caption{{Comparisons of R-DAD without (top) /with (bottom) the region decomposition assembly method under occlusions on MSCOCO 2018 dataset. }}
\vspace{-10pt}
\label{fig:Comp}
\end{center}
\end{figure}

In Fig. \ref{fig:Comp}, we have directly compared detection results with/without the RDA network. Some detection failures (inaccurate localizations  and false positives) for occluded objects are occurred when not using the proposed network.

\begin{table*}[!t] \vspace{-0pt}
\caption{Comparison of state-of-the-art detectors on MSCOCO18 test-dev set. More results can be founded \href{https://competitions.codalab.org/competitions/5181\#results}{in the MSCOCO evaluation website (test-dev2018)}. For each metric, the best results are marked with {\color{red}\underline{\textbf{red}}}. }  
 \vspace{-10pt}
{\scriptsize
\begin{center}
\renewcommand{\tabcolsep}{1.2mm}
\resizebox{\textwidth}{!}{%
\begin{tabular}{c|c|c|ccc|ccc|ccc|ccc}
\hline \hline
 \multirow{2}{*} {\textbf{Detector}}&  \multirow{2}{*}{\textbf{Base Network}} &  \multirow{2}{*}{\textbf{Bells and whistles}} &\multicolumn{3}{c}\textbf{Avg. Precision, IoU:}  & \multicolumn{3}{c}\textbf{Avg. Precision, Area:} & \multicolumn{3}{c}\textbf{Avg. Recall, \# Dets:} & \multicolumn{3}{c}\textbf{Avg. Recall, Area:} \\
& &    & \textbf{0.5:0.95} & \textbf{0.5} &\textbf{0.75}  &\textbf{S} & \textbf{M} & \textbf{L} &\textbf{1}  &\textbf{10}  &\textbf{100} &\textbf{S} &\textbf{M} &\textbf{L} \\ \hline
\multicolumn{15}{c} {\textbf{Single-stage-based dectectors}} \\ 
YOLOv2 \cite{RedmonF_CVPR17}	&DarkNet-19	& - 	&21.6	&44.0	&19.2	&5.0	 &22.4	&35.5	&20.7	&31.6	&33.3	&9.8 &	36.5	&54.4 \\
SSD512 \cite{LiuAESRFB_ECCV16} &VGG-16 &-	&28.8	&48.5	&30.3	&10.9	&31.8	&43.5	&26.1	&39.5	&42.0	&16.5	&46.6	&60.8 \\
DSSD513 \cite{FuLRTB_Corr17}	&ResNet-101	&-	&33.2	&53.3	&35.2	&13.0	&35.4	&51.1	&28.9	&43.5	&46.2	&21.8	&49.1	&66.4\\
RectinaNet \cite{LinGGHD_ICCV17}	&ResNet-101	& - 	&34.4&	53.1&	36.8&	14.7&	38.5&	49.1&- &- &- &- &- \\
RectinaNet	 \cite{LinGGHD_ICCV17}	 &ResNet-101	& Feature pyramid	&39.1&	59.1&	42.3&	21.8&	42.7&	50.2& - &- &- &- &- &- \\
RefineDet512 \cite{Shifeng_CVPR18}	&ResNet-101	& -	&36.4	&57.5	&39.5	&16.6	&39.9	&51.4	&30.6	&49.0	&53.0	&30.0 &58.2	&70.3\\ 
RefineDet512+  \cite{Shifeng_CVPR18}	&ResNet-101	& Multi-scale testing	&41.8&	62.9&	45.7&	25.6&	45.1&	54.1&	34.0&	56.3&	\color{red}\textbf{\underline{\textbf{65.5}}}&\color{red}\textbf{\underline{\textbf{46.2}}}&	\color{red}\textbf{\underline{\textbf{70.2}}}&	\color{red}\textbf{\underline{\textbf{79.8}}}\\ 
\hline \hline
\multicolumn{15}{c} {\textbf{Two-stage-based dectectors}} \\ 
Faster RCNN \cite{RenHGS15_NIPS15} &VGG-16 & - &21.9	 & 42.7 & - & - & - &-&-&-&-&-&-&-\\
OHEM++ \cite{ShrivastavaGG_CVPR16}& VGG-16 &- &25.5 &45.9&	26.1&	7.4&	27.7&40.3&31&46.6&48.1 &27.1 &52.2	&63.6\\
R-FCN \cite{DaiLHS_CORR16}& ResNet-101 & - &29.9	&51.9 &-	&10.8	&32 .8 &45.0\\
ION  \cite{BellZBG_CVPR16} & VGG-16& - &33.1	&55.7	&34.6	&14.5	&35.2	&47.2	&28.9	&44.8	&47.4	&25.5	&52.4	&64.3\\
CoupleNet \cite{ZhuZWZWL_ICCV17}& ResNet-101 & Multi-scale training &34.4 &54.8 &37.2 &13.4	&38.1	&50.8	&30.0	&45.0	&46.4	&20.7	&53.1	&68.5 \\
Faster R-CNN+++ \cite{HeZRS_CVPR16}& ResNet-101-C4	& Multi-scale testing	&34.9	&55.7	&37.4	&15.6	&38.7	&50.9 & - &- &- &- &-&-\\
Feature pyramid network  \cite{LinDGHHB_Corr16}	& ResNet101	& -	&36.2	&59.1	&39.0	&18.2	&39.0	&48.2	&31.0	&46.6	&48.1	&27.1	&52.2	&63.6\\
Deformable R-FCN \cite{jifeng_iccv17}	 &Aligned-Inception-ResNet	&- &36.1&	56.7& - &14.8	& 39.8	& 52.2 & - &- &- &- &-&- 
\\
Deformable R-FCN \cite{jifeng_iccv17}	 &Aligned-Inception-ResNet	& Multi-scale testing&37.1& 57.3& - &18.8	& 39.7	& 52.3 & - &- &- &- &-&-
\\
G-RMI 	\cite{HuangRSZKFFWSG_Corr16}&Inception-ResNet-v2 & - &34.7	& 55.5	&36.7	&13.5	&38.1	&52.0 & - &- &- &- &-&-\\
\textbf{R-DAD (ours)}	&\textbf{ResNet-50}	&- &37.1&	57.7 &	39.9 &	19.6&	 41.2&	52.1&	31.4& 	48.2& 	50.2&	 29.3&	54.7& 	68.1 \\
\textbf{R-DAD (ours)}	&\textbf{ResNet-101}	&- &40.4 &	60.5&	43.7&	20.4&	45.0&	56.1&	32.5&	53.2&	56.9&	34.1&	61.9&	75.2\\
\textbf{R-DAD-v2 (ours)}& \textbf{ResNet-50}&  Multi-scale testing&41.8&	62.7&	46.3&	23.3&	44.6&	53.5&	33.2&	55.3&	59.4&	37.5&	63.0&	75.3\\
\textbf{R-DAD-v2 (ours)}& \textbf{ResNet-101}&  Multi-scale testing&\color{red}\textbf{\underline{\textbf{43.1}}}&	\color{red}\textbf{\underline{\textbf{63.5}}}&	\color{red}\textbf{\underline{\textbf{47.4}}}&	24.1&	\color{red}\textbf{\underline{\textbf{45.9}}}&	\color{red}\textbf{\underline{\textbf{54.7}}}&	\color{red}\textbf{\underline{\textbf{34.1}}}&	\color{red}\textbf{\underline{\textbf{56.7}}}&	60.9&	39.3&	64.3&	76.2\\
\hline \hline
\end{tabular}}
\vspace{-10pt}
\end{center}}
\label{TABLE:COCO}
\end{table*}

\vspace{0pt}
\section{Conclusion}
In this paper, we have proposed a region decomposition and assembly detector to solve a large scale object detection problem. We first decompose a whole object region into multiple small regions, and learn high-level semantic features by combining a holistic and part model features stage-by-stage using the proposed method. For improving region proposal accuracy, we generate region proposals of various sizes by using our multi-scale \SH{region proposal} method and extract wrapped CNN features within the generated proposals for capturing local \SH{details} of an object and global context cues around \SH{the} object. From \SH{the} extensive comparison with other state-of-the-art convolutional detectors, we have proved that the proposed methods lead to the noticeable performance improvements on several  benchmark challenges such as PASCAL VOC07/12, and MSCOCO18. We clearly  show \SH{that} the  robustness and \SH{flexibility} of our methods by implementing several versions of R-DADs with different feature extractors and detection methods through ablation studies.

\section{Acknowledgement}
This work was supported by the National Research Foundation of Korea (NRF) grant funded by the Korea government (MSIT) (No. NRF-2018R1C1B6003785).

\bibliographystyle{aaai}
\footnotesize
\bibliography{RDAD_aaai_fn}

\end{document}